\documentclass[10pt,twocolumn,letterpaper]{article}

\usepackage[pagenumbers]{cvpr}        %

\usepackage{graphicx}
\usepackage{multirow}
\usepackage{booktabs}
\usepackage{wrapfig}
\usepackage{xspace}
\usepackage{amsmath}
\usepackage{amssymb}
\usepackage{cuted}
\usepackage{capt-of}
\newcommand{\ours}{\textsc{ReconPlusGen}\xspace}

\definecolor{cvprblue}{rgb}{0.21,0.49,0.74}
\usepackage[pagebackref,breaklinks,colorlinks,allcolors=cvprblue]{hyperref}

\title{ReconPlusGen: Injecting Reconstruction Prior into Multi-view 3D Generation through Noise Inversion and Modulation}

\author{
Jiarui Liu\textsuperscript{1}\textsuperscript{*}\qquad
Heng Li\textsuperscript{1}\textsuperscript{*}\qquad 
Weiyu Li\textsuperscript{1}\textsuperscript{2}\qquad 
Keng Deng\textsuperscript{1}\qquad  
Junyuan Deng\textsuperscript{1}\qquad 
Zheng Zhongxing\textsuperscript{3}\qquad \\ 
Junyu Huang\textsuperscript{3}\qquad
Jiahao Chang\textsuperscript{4}\qquad   
Xiaoguang Han\textsuperscript{4}\qquad  
Ping Tan\textsuperscript{1}\textsuperscript{\ensuremath{\dagger}}\qquad  
\\
\textsuperscript{1}HKUST\qquad \textsuperscript{2}LightIllusions\qquad \textsuperscript{3}BYD\qquad   \textsuperscript{4}CUHK-Shenzhen \\
{\small \textsuperscript{*}Core contributions\quad
\textsuperscript{\ensuremath{\dagger}}Corresponding author.}
}

\begin{document}
\maketitle

\begin{strip}
    \centering
    \includegraphics[width=0.95\textwidth]{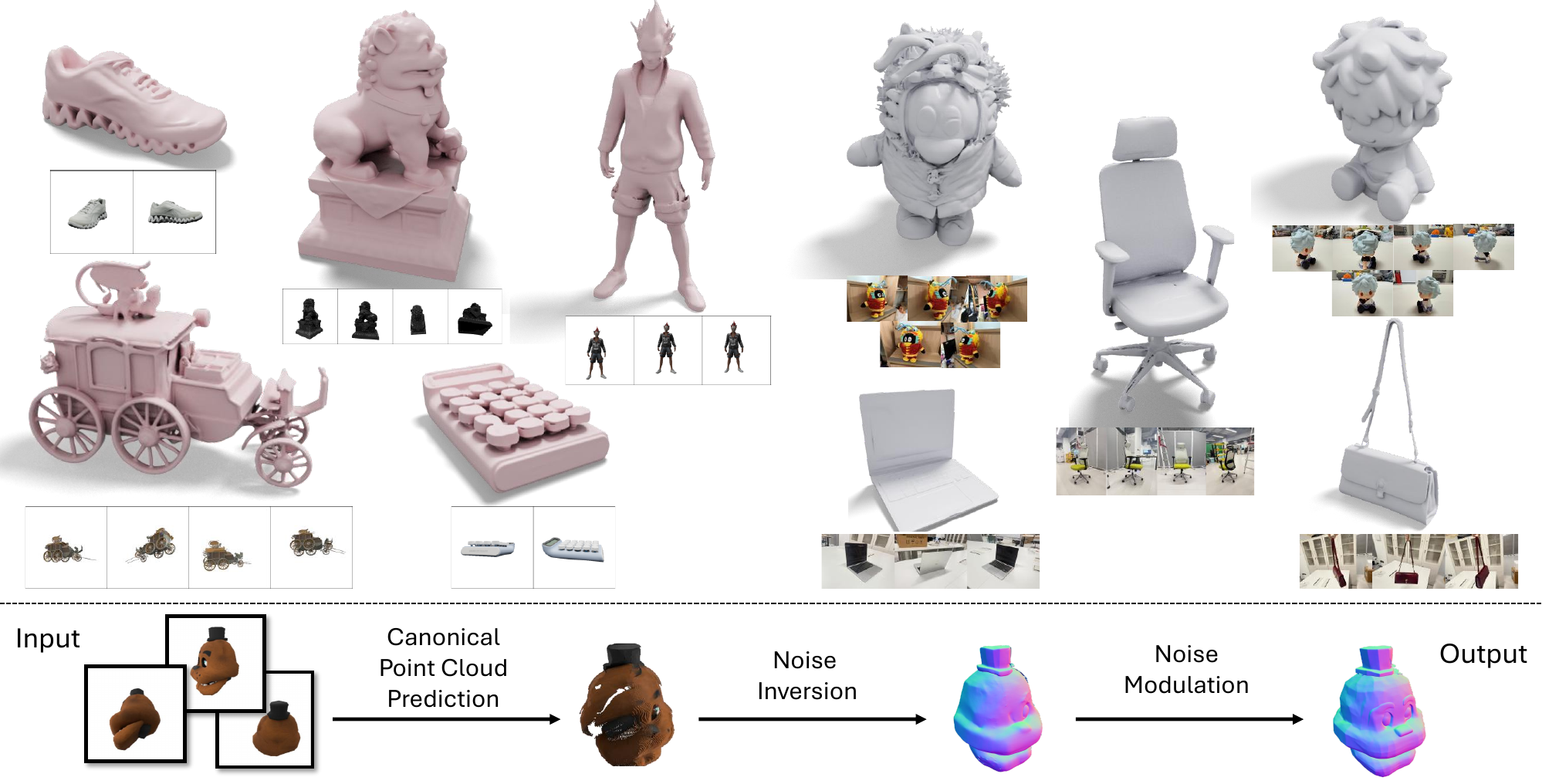}
    \captionof{figure}{Qualitative results and an illustration of our core idea. Top left: reconstruction results on benchmark images. Top right: reconstruction results on real-world images. Bottom: illustration of reconstruction-guided noise initialization and modulation. Given multiple input images, we predict a point cloud in canonical space, deterministically inject the predicted geometry into the diffusion process through noise inversion, and modulate the resulting noise to preserve the generative flexibility required to complete unobserved regions and refine visible geometry.}
    \vspace*{25pt}
    \label{fig:teaser}
\end{strip}

\begin{abstract}
Multi-view 3D reconstruction and 3D generation offer complementary properties: reconstruction models preserve observation-grounded geometry but often fail to recover unobserved regions, whereas generative models synthesize plausible 3D assets but struggle to maintain pixel-level geometric fidelity because of the stochastic nature of diffusion. Existing methods attempt to combine the strengths by conditioning generative models on reconstruction features but still treat both observed and unobserved regions stochastically.
We present \ours, a unified framework that deterministically injects reconstruction priors into a 3D generative model through its \emph{initial-noise space}. Given a set of unposed images, our approach reconstructs the visible geometry and transforms it into a structured diffusion initialization via noise inversion. To adaptively balance reconstruction fidelity and generative flexibility, we introduce a confidence-guided, spatially varying noise modulation scheme that strongly constrains reliable regions, relaxes constraints in uncertain regions, and preserves generative freedom in unobserved regions. Finally, a multi-view-conditioned diffusion model refines geometric details while maintaining consistency with the input observations. Extensive experiments on synthetic and real-world benchmarks demonstrate that \ours consistently outperforms existing reconstruction-guided and generation-based approaches. These results establish initial-noise control as a simple yet effective mechanism for faithful, high-quality 3D asset generation from multi-view observations.
\end{abstract}

\vspace*{-20pt}
\section{Introduction}
\label{sec:intro}
3D object reconstruction has long been a fundamental problem in 3D computer vision, with broad applications in VR/AR and content creation. Conventional reconstruction methods~\cite{pan2024glomap,snavely2006photo,schoenberger2016sfm} rely on reliable cross-view correspondences to recover the 3D structure. Such pipelines are brittle in low-texture regions, and need multi-stage processing or iterative optimization~\cite{guedon2025matcha,DP-GS}.
Recent learning-based methods have made strong progress in sparse-view 3D object reconstruction. Regression models~\cite{wang2025vggt,deng2025sailrecon,yang2025fast3r,depthanythingv3, wang2026vggtomega} infer 3D attributes from unposed image sets, improving performance under challenging imaging conditions. Nevertheless, these approaches are inherently observation-bounded: they can only predict 3D attributes for observed pixels, resulting in incomplete reconstructions.

Recent advances in diffusion-based 3D generative models~\cite{zhang20233dshape2vecset,zhang2024clay,hunyuan3d2025hunyuan3domni,hunyuan3dv21,li2024craftsman3d} offer a compelling way to predict complete shapes from limited observations. By learning strong 3D priors from large-scale 3D data, these methods can generate complete 3D content conditioned on sparse-view images. Such generative priors can synthesize unobserved regions with high-quality geometry and appearance, improving reconstruction completeness by filling in missing structures and details. Some works~\cite{chang2025reconviagen,Ultra3D} attempt to improve the pixel-level reconstruction alignment by replacing 2D image features with 3D point cloud features.
However, diffusion inference conditioned on features is inherently stochastic, which introduces substantial uncertainty and sample-to-sample variability. This variability makes it difficult to achieve the pixel-level and multi-view alignment needed for precise geometric consistency.

Instead of conditioning the denoising process on geometry features, we take a different perspective: our geometry-grounded generation framework, \ours, injects deterministic reconstruction cues \emph{at the source of stochasticity}---the initial noise of the generative process.
Specifically, we first train a VGGT-style geometry predictor, termed Canonical-Aligned VGGT (CA-VGGT), to estimate geometry directly in a shared canonical space from unposed observations, providing an initial shape for noise inversion. We then apply confidence-aware spatial noise modulation to inversion noise to enforce geometric fidelity in reliable regions, relax constraints in uncertain regions, and preserve generative flexibility in unobserved regions. Finally, a multi-view diffusion recovers fine-grained geometric details while maintaining global and cross-view consistency from the modulated initial noise. This design yields a controllable yet expressive generation process that improves both the accuracy of the reconstruction and the perceptual quality.

Extensive experiments on synthetic and real-world image benchmarks demonstrate that \ours achieves state-of-the-art performance in unposed sparse-view settings. Compared with prior feature-conditioned pipelines, our method provides stronger geometric fidelity, improved image alignment, and more robust completion of unseen regions. 
Our contributions are summarized as follows:
\noindent
\begin{itemize}
    \item We introduce a novel paradigm for geometry injection in 3D diffusion models: reconstructing visible geometry in an object-centric canonical space and inverting it into a corresponding \emph{initial noise} to provide a deterministic anchor for the diffusion trajectory.
    \item We propose a confidence-aware, spatially varying noise modulation scheme that adaptively controls the influence of the reconstructed geometry according to its estimated reliability, preserving fidelity in confident regions while retaining generative flexibility in uncertain and unobserved regions.
    \item We achieve state-of-the-art performance across diverse sparse-view 3D object reconstruction benchmarks.
\end{itemize}

\begin{figure*}[tb]
    \centering
    \includegraphics[width=0.95\textwidth]{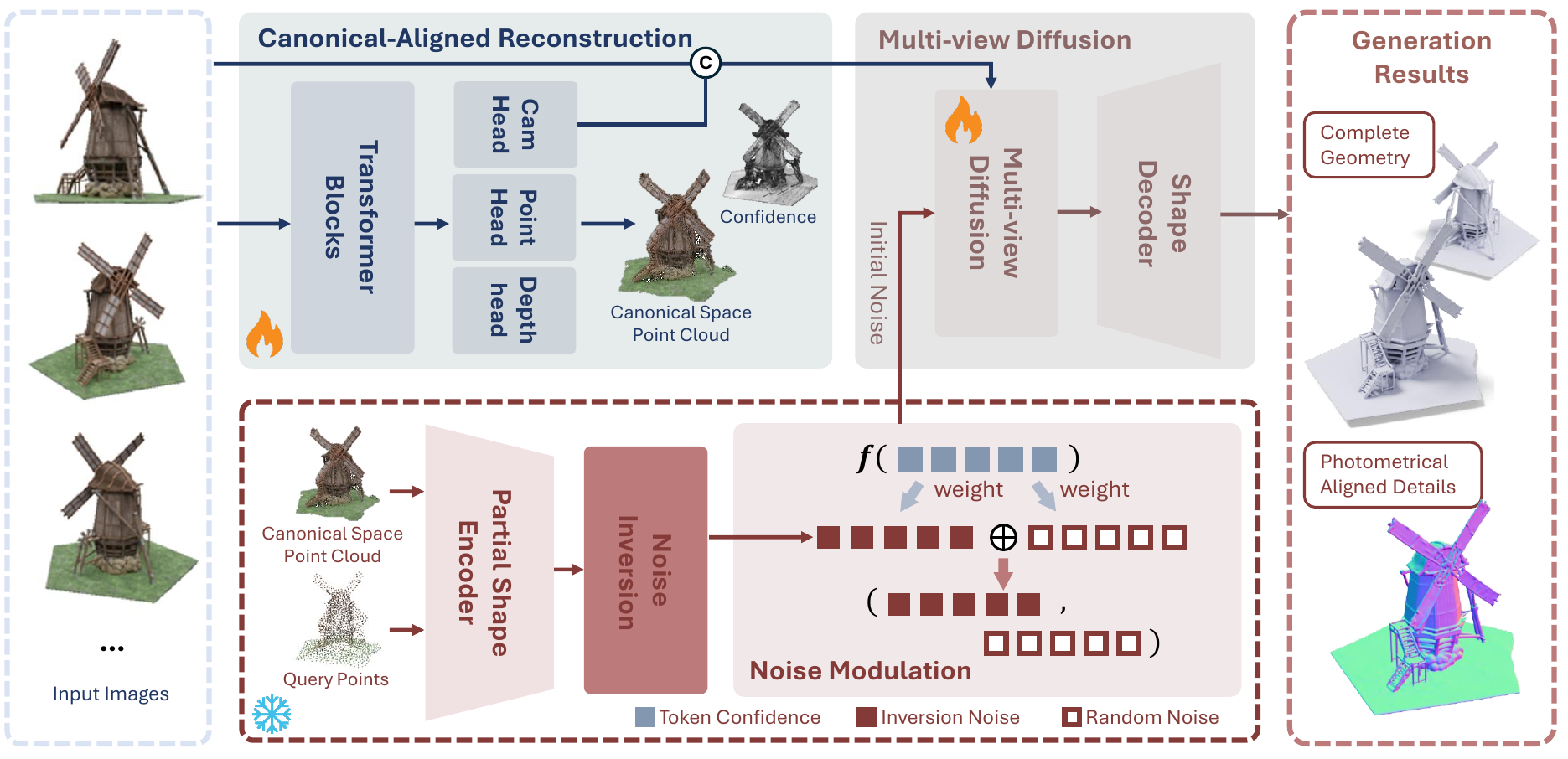}
    \caption{Pipeline of \ours. Given multiple input images, our canonical-aligned geometry prediction model first reconstructs the visible geometry directly in canonical space. We then inject the predicted geometry into the diffusion process through noise inversion and modulation. Starting from the modulated noise, a multi-view image-conditioned diffusion model generates the final shape while preserving observed geometry and completing unobserved regions.}
    \label{fig:pipeline}
\end{figure*}

\section{Related Works}

\subsection{3D Reconstruction}
Traditional Optimization-based reconstruction methods~\cite{yariv2021neuS,zhang2024radegs,Zhang2026GeometryGrounded} can recover highly accurate geometry, but typically rely on dense observations and known camera parameters.
Recent approaches~\cite{guedon2025matcha,DP-GS} extend this paradigm to sparse-view settings by incorporating priors from pretrained geometry models. Nevertheless, their costly per-scene optimization still limits both computational efficiency and applicability to downstream tasks.
Recent feed-forward models~\cite{wang2024dust3r,duisterhof2024mast3r} directly predict depth maps and camera parameters from unposed images. Subsequent approaches~\cite{deng2025vggtlong,maggio2025vggtslam,deng2025sailrecon,wang2026vggtomega} improve scalability to multi-view inputs and even support large-scale reconstruction from thousands of images. Despite their strong performance, these models remain observation-bounded: they primarily recover geometry corresponding to visible pixels and do not explicitly address complete object reconstruction.

Different from scene-level reconstruction methods above,
object-centric reconstruction methods instead model object shapes in a normalized canonical space, usually defined in $[-1,1]$. LucidFusion~\cite{he2025lucidfusionreconstructing3dgaussians} proposes to predict object geometry in canonical space coordinates from sparse-view images. Large reconstruction models~\cite{hong2023lrm,li2023instant3d,tang2024lgm,xu2023dmv3d,xu2024instantmesh,wei2025meshlrmlargereconstructionmodel} further focus on recovering complete geometries. However, their outputs are often overly smooth and fail to preserve fine geometric details.

\subsection{3D Generative Models}
Large-scale 3D generative models provide a complementary direction by learning strong shape and appearance priors from extensive 3D datasets.
Shape2VecSet~\cite{zhang20233dshape2vecset} encodes signed distance fields into compact sets of latent tokens, enabling diffusion models to operate directly in this compressed representation space. Building on this paradigm, Clay~\cite{zhang2024clay} generates complete textured 3D meshes from a single image. Subsequent works~\cite{li2024craftsman3d,trellis,zhu2023hifa,hunyuan3dv20,detailgen3d,li2025sparc3d} further improve generation quality through advanced architectures and sparse representations. 
However, most of these methods are conditioned on a single image or assume restrictive camera configurations~\cite{hunyuan3dv20}, and their generated assets may deviate from the observed geometry.

To support arbitrary unposed input images, previous approaches bridge reconstruction and generation by conditioning the denoising process on geometry-aware features~\cite{chang2025reconviagen}. Some of them improve geometric consistency by finding 2D-3D correspondences~\cite{Ultra3D,CUPID,li2026pixal3d}, others~\cite{huang2026recgen3d,lin2026mix3rmixingfeedforwardreconstruction} explicitly learn the alignment between reconstruction and generation representations.
Although these designs improve controllability, they still operate mainly at feature condition level. Therefore, structural sampling can remain unstable due to stochastic diffusion dynamics. 
Our method differs by injecting reconstruction priors into the initial noise, which provides stronger geometric control with less reliance on strict calibration.

\subsection{Initial Noise in Diffusion Models.}
Recent studies on diffusion models show that the initial noise strongly affects the generation result.
Pioneer works~\cite{Priorgrad,GoodSeed} found that the noise space itself contains structural cues, motivating research in test-time optimization, learned noise priors, and diffusion inversion. Test-time methods~\cite{DOODL,InitNO,D-Flow} use gradient-based optimization to refine the initial noise for better alignment, while Reno~\cite{Reno} extends this idea to one-step models with multiple reward signals. However, these methods introduce additional computation during inference. To reduce this cost, several approaches~\cite{Noisehypernetworks,Goldennoise,NoiseRefine} train lightweight networks to predict improved noise in a single pass. DeepInv~\cite{DeepInv} uses a self-supervised trainable solver for faster image-to-noise mapping. NoiseAR~\cite{NoiseAR} instead learns an autoregressive noise prior for controllable generation and reinforcement learning. Meanwhile, diffusion inversion methods map an image back to its initial noise.
This has also become a common foundation for image editing~\cite{NullTextInversion,Edict,Renoise}. Recent works further improve inversion fidelity, editability, and efficiency through different schedules~\cite{kang2024etainversion,huang2025dualschedule}, high-order solvers for rectified-flow models~\cite{wang2025RFSolver}, and learned one-step inversion~\cite{nguyen2025swiftedit}.
Together, these studies show that controlling the initial noise can improve both generation quality and controllability. 
Inspired by this finding, we extend initial-noise control to reconstruction-guided 3D generation by converting predicted geometry into structured noise.

\section{Method}
We present \ours, a geometry-grounded 3D object generation framework that takes a set of unposed input images and produces a high-fidelity, pixel-aligned, and multi-view-consistent 3D mesh.
Given a set of unposed images, we first recover camera poses and reconstruct a point cloud in the canonical space via Canonical-Aligned VGGT. Building upon this geometric prior, we describe in Sec.~\ref{sec:method:inversion} how to obtain a deterministic initial noise from the reconstructed point cloud. Starting from this geometry-informed initialization, we further modulate the initial noise in Sec.~\ref{sec:method:confnoisemod} to facilitate the recovery of geometry in unobserved regions. Finally, Sec.~\ref{sec:method:diffusion} details the diffusion process conditioned on multi-view inputs with camera embeddings.
We adopt the 3DShape2VecSet~\cite{zhang20233dshape2vecset} based 3D generation framework~\cite{hunyuan3dv21} as our backbone.

\subsection{Noise Initialization with Reconstruction}
\label{sec:method:inversion}

\begin{figure}[!t] %
    \centering
    \includegraphics[width=1.0\linewidth]{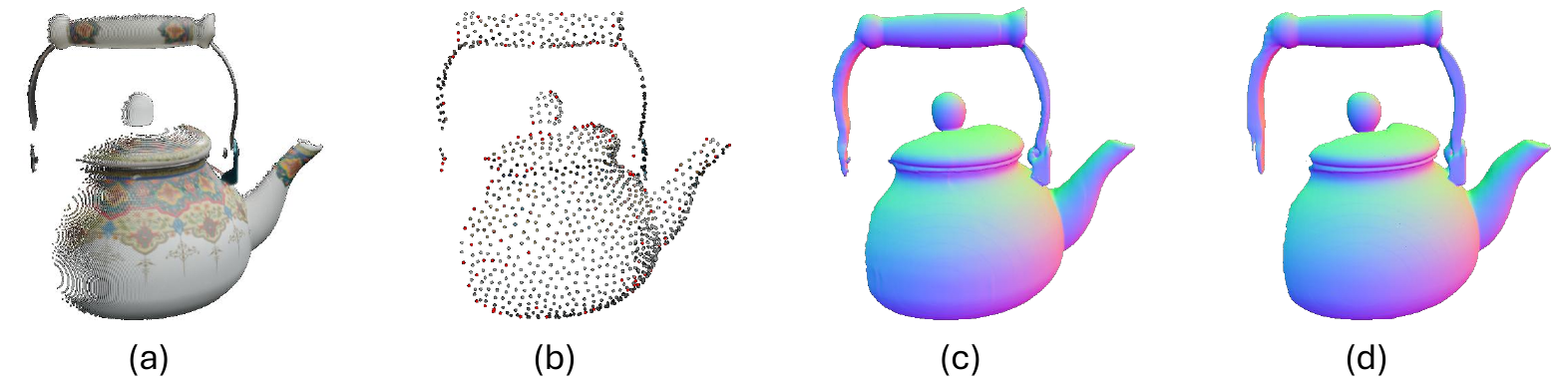} %
    \caption{Spatial Locality of 3DShape2VecSet~\cite{zhang20233dshape2vecset} based diffusion models. (a) Given a partial point cloud, (b) we sample query points $q$ using farthest point sampling~\cite{moenning2003fps} (c) We encode the point cloud into the latent space with $q$, and decode it into a mesh. (d) We perform noise inversion~\cite{wang2025RFSolver} and recover the shape through denoising and decoding.} %
    \label{fig:noise_inversion}
\end{figure}

\paragraph{Deterministic Noise Initialization.}
Starting from a random initial noise, a common practice for 3D generation~\cite{chang2025reconviagen,li2026pixal3d} is to inject feature conditions via cross-attention at each diffusion step. However, such feature conditioning is often insufficient for precise geometric control because the initial state is still unconstrained noise, introducing stochastic variation in both visible and unobserved regions, as shown in Fig.~\ref{fig:noise-continuity}. 
Recent studies have shown that diffusion outputs are strongly determined by initial noise~\cite{NoiseAR, NoiseRefine, Goldennoise, Noisehypernetworks}. 
This observation motivates us to inject the condition in an alternative manner: recovering the corresponding deterministic noise from a given explicit geometry prior.

Given a reconstructed point cloud $S$, we first encode it in the latent representation $Z_{0}=\text{Encoder}(S,q)$ and then recover the corresponding initial noise $Z^{\mathrm{inv}}=Inv(Z_{0})$ through noise inversion, denoted by $Inv(\cdot)$. Here, the query points $q$ are selected from $S$ using farthest point sampling~\cite{moenning2003fps} method and provided as input to the shape encoder. We adapt the inversion scheme of~\cite{wang2025RFSolver} to native 3D diffusion models, approximating the velocity term through a Taylor expansion:
\begin{equation}
Z_{t_{i+1}} = Z_{t_i} + \sum^{n - 1}_{k=0} \frac{((t_{i+1} - t_i)^{k+1})}{(k + 1)!} 
v^{(k)}_{\theta}(Z_{t_{i}}, t_i)+ \mathcal{O} (h^{n+1}_i),
    \label{equ:RF-Solver}
\end{equation}
where $Z_{t_{i+1}}$ denotes the latent vector at timestep $i$, $v^{(k)}_{\theta}(\cdot)$ denotes the $k$-th derivative of the velocity, $v^{(0)}_{\theta}(\cdot)$ is predicted by flow-matching model $\mathcal{D}_{\theta}$. $\mathcal{O}(h_i^{n+1})$ is the Peano remainder which is omitted in the computation. $n$ denotes the Taylor expansion order, where we empirically set the expansion order at $n=2$.
We approximate the higher-order terms using finite differences:
\begin{equation}
    v_{\theta}^{(k+1)}(Z_{t},t) \approx \frac{v_{\theta}^{(k)}(Z_t,t) - v_{\theta}^{(k)}(Z_{t+\Delta t},t+\Delta t)}{\Delta t}
\end{equation}
We visualize the input point cloud $S$, the mesh reconstructed by the decoder $M=\text{Decoder}(Z_0)$, and the mesh generated from the inverted noise $M=\text{Decoder}(\text{DiT}(Z_T))$ in Fig.~\ref{fig:noise_inversion}. The results demonstrate that the inverted noise initialization can faithfully preserve the given geometry prior: the reconstructed region aligns well with the input point cloud, while the unobserved regions remain empty.

\paragraph{Canonical Point Cloud Generation.}
Another challenge in noise inversion lies in the misalignment between the coordinate system of the input reconstructed point cloud and the canonical space required by the diffusion network. The point cloud from the estimated method is usually defined up to a scale~\cite{schoenberger2016sfm} or with respect to a specific frame coordinate~\cite{wang2025vggt}. Directly using such point clouds for noise inversion leads to a catastrophic failure of diffusion prediction due to the discrepancy in the input domains, as noted in RecGen3D~\cite{huang2026recgen3d}.
To address this misalignment of coordinate system,
we train a \textbf{Canonical-Aligned VGGT (CA-VGGT)} to estimate geometry from unposed images in canonical space. Instead of assuming the camera pose of the first frames as identity, we train CA-VGGT directly in the standard canonical coordinate range of $[-1,1]$.
We predict point clouds and camera poses with different heads, as well as the corresponding confidence maps, denoted as $C$. Those confidence maps are designed to be proportional to the model's prediction error~\cite{wang2025vggt} and to provide an indicator of reconstruction quality in Sec~\ref{sec:method:confnoisemod}. Please refer to the supplementary material for more details about CA-VGGT.

As a result, for a given set of input images, our method identifies a deterministic diffusion initialization that produces a shape well aligned with the observation, eliminating frame-dependent position, scale, and pose ambiguity and reducing stochastic variation in visible regions, thereby substantially improving reconstruction-to-generation consistency.

\subsection{Confidence-aware Noise Modulation}
\label{sec:method:confnoisemod}

\begin{figure}[!t] %
    \centering
    \includegraphics[width=1.0\linewidth]{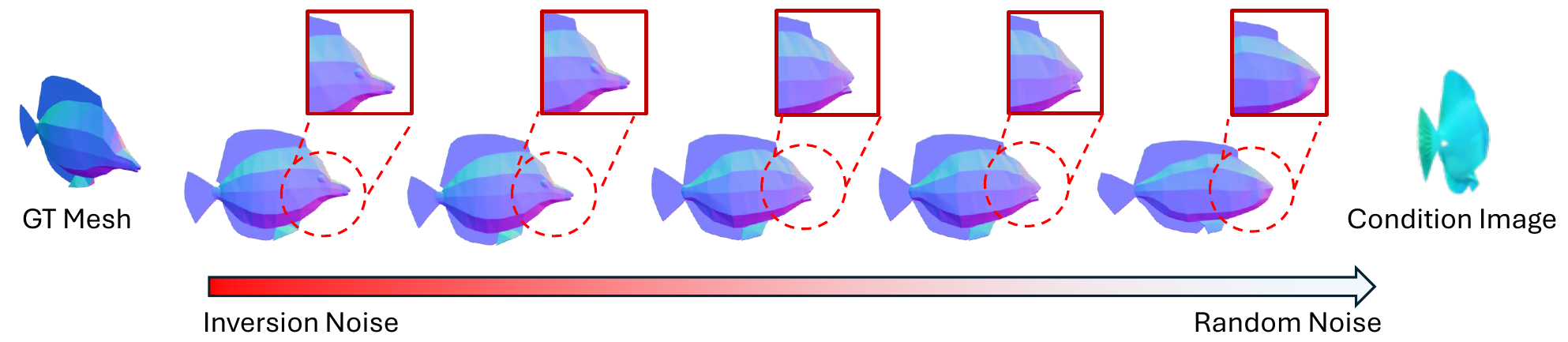} 
    \caption{Continuity of noise space in a 3DShape2VecSet~\cite{zhang20233dshape2vecset} based diffusion model. 
    We compute inversion noise from ground truth mesh(GT mesh), interpolate between the resulting inversion noise and a random noise, and conduct denoising starting from each interpolated noise. Both inversion and denoising process is conditioned with the same image(shown in rightmost).} %
    \label{fig:noise-continuity}
\end{figure}

\label{sec:method:modulation}
Sec.~\ref{sec:method:inversion} provides a deterministic initialization that anchors the diffusion process to the given geometry prior.
However, the point cloud estimated from the images is inherently incomplete and contains prediction errors, which leads to inferior quality of the generated mesh. 
To address these limitations, we introduce a \emph{spatially varying, confidence-aware} noise modulation scheme that preserves reliable geometric cues while relaxing constraints in uncertain and unobserved regions, with respect to two key properties of the 3DShape2VecSet~\cite{zhang20233dshape2vecset} based diffusion models: \textbf{Spatial Locality} and \textbf{Noise Initialization Continuity}.

\paragraph{Spatial Locality.} As shown in Fig.~\ref{fig:noise_inversion}, each query point in $q$ constitutes a latent token that controls a spatially localized region during encoding. Regions devoid of query points cannot be adequately represented by latent tokens. This property remains after noise inversion.
Consequently, we selectively employ a subset of query points (tokens) $\tilde{q}\subset q$ to encode the observed portion of the point cloud, perform noise inversion, while leaving the remaining tokens as random noise. This strategy, termed \emph{noise padding} (Noise-Pad), enables the model to synthesize geometry in unseen regions from scratch:
\begin{equation}
Z_{T} = [Z^{\mathrm{inv}}=inv\bigl(\text{Encoder}(S,\tilde{q})\bigr),\; Z^{\mathrm{random}}].
\label{equ:noise_padding}
\end{equation}
Note that determining the completeness ratio of the input point cloud is non-trivial. For simplicity, we use $2,048$ query points and $2,048$ random tokens, forming a total of $4,096$ latent tokens for the 3DShape2VecSet VAE~\cite{hunyuan3dv21}.

\paragraph{Noise Initialization Continuity.}
Although deterministic noise initialization on the given points provides a strong prior for 3D generation, a noisy or inaccurate prior can lead to inferior results. Note that both $Z^{\mathrm{inv}}$ and $Z^{\mathrm{random}}$ are drawn from $\mathcal{N}(0, I)$. By the closure property of Gaussian distributions under linear combinations, any convex combination of them remains Gaussian. This motivates us to interpolate between $Z^{\mathrm{inv}}$ and $Z^{\mathrm{random}}$ according to the reconstruction quality of each token.
The confidence map $C$ predicted by CA-VGGT provides a natural basis for determining the interpolation weights: it reflects the reliability of the reconstructed geometry and is spatially aligned with each point in the reconstructed point cloud $S$. We therefore use these confidence scores to modulate the geometric constraints as follows:

\begin{equation}
Z^{\mathrm{recon}} = \sqrt{\alpha}\cdot Z^{\mathrm{inv}}
+ \sqrt{(1-\alpha)} \cdot Z^{\mathrm{random}},
\end{equation}
where $\alpha=f(C(\tilde{q}))$, $C(\tilde{q})$ denotes the confidence scores associated with the selected query points, and $f(\cdot)$ is a linear function that maps confidence values to interpolation weights: $f(c)=c^{min}+(1-c^{max})\cdot n(c)$. Here, $ n(c)\in[0,1]$ denotes the normalized value of $c$.
We visualize the effect of interpolating $Z^{\mathrm{recon}}$ in Fig.~\ref{fig:noise-continuity}. Starting from each interpolated noise, we perform denoising under the same image conditioning and decode the resulting latent representation into a mesh. As the initialization shifts from inversion noise toward random noise, the generated shapes change smoothly, demonstrating the continuity of the noise space. This transition gradually transfers geometric control from the reconstruction prior encoded in the inversion noise to the learned generative prior. Consequently, high-confidence tokens remain close to the inversion noise, whereas low-confidence tokens contain more random noise and can be refined more freely by the generative model.
Finally, we replace $Z^{\mathrm{inv}}$ in Eq.~\ref{equ:noise_padding} with $Z^{\mathrm{recon}}$, yielding the following modulated noise tokens:
\begin{equation}
\label{equ:final_token}
Z_{T} = [Z^{\mathrm{recon}}, {Z}^{\mathrm{random}}].
\end{equation}

Overall, the modulated tokens preserve the reconstructed geometry in reliable regions while delegating the refinement of uncertain regions and the completion of unseen regions to the learned diffusion prior, thereby achieving a favorable balance between input alignment and geometric quality.

\subsection{Multi-View Diffusion Refinement}
\label{sec:method:diffusion}
Starting from the noise in Eq.~\ref{equ:final_token}, the original single-view diffusion model without fine-tuning is already sufficient to generate the complete mesh, making our method entirely training-free. However, a multi-view diffusion model remains indispensable for complementing the reconstruction by correcting residual errors and enriching fine-scale geometric details. To this end, we extend the single-view image conditioning to a multi-view setting by incorporating camera pose embeddings. Specifically, we convert the camera poses $\tilde{\mathbf{P}}$ estimated by CA-VGGT into view-specific Pl\"ucker ray maps as explicit pose embeddings. These pose embeddings are added to the DINO image features channel-wise. We then concatenate the conditioning tokens across views and feed them into the diffusion model:
\begin{equation}
V = \mathcal{D}_{\theta}\left(Z; \text{DINO}(\mathbf{I}) \oplus \text{Pl\"uckerEmbed}(\tilde{\mathbf{P}})\right),
\end{equation}
where $\mathcal{D}_{\theta}$ denotes the multi-view diffusion model, $Z$ is the latent token set at timestep $t$, $\mathbf{I}$ denotes the input RGB images, and $\tilde{\mathbf{P}}$ denotes the corresponding camera poses. The operator $\oplus$ indicates channel-wise addition. This design allows the diffusion model to exploit both semantic context and explicit geometric cues, facilitating the recovery of fine-grained surface details while maintaining multi-view consistency.

\section{Experiments}
\subsection{Implementation Details}
\paragraph{Training.} We select 130K high-quality meshes from Objaverse~\cite{deitke2022objaverseuniverseannotated3d} as training set and render 60 random views per mesh with camera jittering. This dataset is used for training both CA-VGGT and the multi-view diffusion model. All meshes are normalized so that their vertex coordinates fall within $[-1, 1]$.
We initialize CA-VGGT with the pretrained weights of VGGT~\cite{wang2025vggt}. During training, we dynamically sample one to eight input views for each batch. Within each batch, we randomly select one image as the first frame defining the canonical axes and transform the camera poses and point clouds of all views into the corresponding canonical coordinate system. We train CA-VGGT on 16 H100 GPUs for 17k iterations using a learning rate of $1\times10^{-5}$.
For the multi-view diffusion model, we fine-tune the diffusion model from Hunyuan3D-2.1~\cite{hunyuan3dv21} with the multi-view condition for 110K iterations on 16 H100 GPUs.

\paragraph{Evaluation.} Following ReconViaGen~\cite{chang2025reconviagen}, we randomly sample 300 objects from DoraBench~\cite{Chen_2025_Dora}, 200 objects from OmniObject3D~\cite{wu2023omniobject3d} and 300 objects from Objaverse~\cite{deitke2022objaverseuniverseannotated3d}. For DoraBench, we render 24 images per object using the TRELLIS~\cite{trellis} rendering protocol. 
For OmniObject3D, we randomly select four views from the 24-view Blender renderings contained in the original dataset. 
We evaluate on a held-out set of meshes sampled from Objaverse, ensuring no overlap with the training data, and render them using the same configuration as the training set.
We adopt the $\ell_1$ Chamfer distance and the F-score at a threshold of $0.1$ as evaluation metrics. Before computing these metrics, we rescale each prediction and align it with the ground-truth shape using a scaled ICP initialized from four different orientations, thus mitigating the scale and orientation ambiguities introduced by different reconstruction models.

\subsection{Quantitative Results}

\begin{table*}[t]
\centering
\caption{Quantitative comparison on three benchmarks with the 4-view setting. We report Chamfer Distance (CD, $\downarrow$) and F-Score ($\uparrow$, threshold=0.1). Best results under the standard setting are shown in \textbf{bold}. Results shown in gray report the empirical upper bound obtained using ground-truth point clouds. $\checkmark$ indicates methods that require ground-truth camera poses, and $^{*}$ indicates methods conditioned on point clouds predicted by CA-VGGT.}
\resizebox{0.75\textwidth}{!}{  %
\begin{tabular}{cl|cc|cc|cc} 
\toprule
\multirow{2}{*}{Type} & \multirow{2}{*}{Method} & \multicolumn{2}{c|}{Objaverse} & \multicolumn{2}{c|}{Dorabench} & \multicolumn{2}{c}{OmniObject3D} \\
\cline{3-8}
 & & CD $\downarrow$ & F-Score $\uparrow$ & CD $\downarrow$ & F-Score $\uparrow$ & CD $\downarrow$ & F-Score $\uparrow$ \\
\midrule
\multirow{4}{*}{\begin{tabular}[c]{@{}c@{}}Reconstruction \\ Methods\end{tabular}} 
 & LGM~\cite{tang2024lgm}~$\checkmark$ & 0.246 & 0.562 & 0.077 & 0.918 & 0.090 & 0.878 \\
 & InstantMesh~\cite{xu2024instantmesh}~$\checkmark$ & 0.242 & 0.581 & 0.094 & 0.874 & 0.071 & 0.934 \\
 & MAtCha GS~\cite{guedon2025matcha}~$\checkmark$ & 0.124 & 0.814 & 0.097 & 0.863 & 0.104 & 0.868  \\
  & DP-GS~\cite{DP-GS}~$\checkmark$ & 0.140 & 0.776 & 0.099 & 0.859 & 0.113 & 0.841  \\
\midrule
\multirow{4}{*}{\begin{tabular}[c]{@{}c@{}}3D Generative \\ Models\end{tabular}} 
 & Craftsman~\cite{li2024craftsman3d} & 0.141 & 0.774 & 0.123 & 0.820 & 0.109 & 0.849 \\
 & Hunyuan2.1~\cite{hunyuan3dv21} & 0.087 & 0.880 & 0.084 & 0.893 & 0.078 & 0.913 \\
 & Hunyuan2.0-MV~\cite{hunyuan3dv20} & 0.085 & 0.890 & 0.092 & 0.87 & 0.075 & 0.924 \\
 & Trellis-MV~\cite{trellis} & 0.088 & 0.880 & 0.065 & 0.941 & 0.070 & 0.940 \\
\midrule
\multirow{3}{*}{\begin{tabular}[c]{@{}c@{}}Point Cloud\\ Conditioned Models\end{tabular}} 
 & HunyuanOmni~\cite{hunyuan3d2025hunyuan3domni}$^{*}$ & 0.077 & 0.905 & 0.076 & 0.905 & 0.059 & 0.945 \\
 & DeepMesh~\cite{zhao2025deepmesh}$^{*}$ & 0.117 & 0.803 & 0.141 & 0.774 & 0.143 & 0.789 \\
 & BPT~\cite{bpt}$^{*}$ & 0.101 & 0.839 & 0.112 & 0.823 & 0.086 & 0.884 \\
\midrule
\multirow{5}{*}{\begin{tabular}[c]{@{}c@{}} Combined Methods \end{tabular}} 
 & ReconViaGen~\cite{chang2025reconviagen} & 0.053 & 0.961 & 0.052 & 0.965 & 0.052 & 0.976 \\
  & Mix3R~\cite{lin2026mix3rmixingfeedforwardreconstruction} & 0.073 & 0.912 & 0.056 & 0.949 & 0.054 & 0.960 \\
  & MV-SAM3D~\cite{li2026mvsam3d} & 0.078 & 0.907 & 0.061 & 0.945 & 0.075 & 0.926 \\
 & \textbf{Ours}$^{*}$ & \textbf{0.040} & \textbf{0.974} & \textbf{0.042} & \textbf{0.966} & \textbf{0.044} & \textbf{0.981} \\
& \textcolor{darkgray}{Ours(GT Point Cloud)} & \textcolor{darkgray}{0.017} & \textcolor{darkgray}{0.990} & \textcolor{darkgray}{0.026} & \textcolor{darkgray}{0.982} & \textcolor{darkgray}{0.020} & \textcolor{darkgray}{0.991} \\
\bottomrule
\end{tabular}
}
\label{tab:results}
\end{table*}

\paragraph{Comparison with Baseline Methods.}
We compare our method with four categories of baselines: reconstruction-based methods, image-conditioned 3D generation models, point-cloud-conditioned generation models, and recent approaches that integrate reconstruction and generation models. We report the results in Tab.~\ref{tab:results} and Tab.~\ref{tab:other_views}.

The reconstruction-based baselines include large feed-forward reconstruction models~\cite{tang2024lgm,xu2024instantmesh} and optimization-based reconstruction methods~\cite{guedon2025matcha, DP-GS}.
Large reconstruction models produce coarse mesh surfaces due to the limited representational capacity of its triplane representation, while optimization-based methods fail to generate reasonable surfaces in unseen regions and result in unsatisfactory indicators even with ground-truth camera poses.

For 3D generative models, we compare against single-view-conditioned approaches~\cite{li2024craftsman3d,hunyuan3dv21} and multi-view-conditioned approaches~\cite{hunyuan3dv20,trellis}. Although these methods achieve marginally better quantitative results, as shown in Fig.~\ref{fig:comparison}, they tend to recover incorrect geometric details due to the lack of an explicit geometric prior.

Point-cloud-conditioned methods~\cite{hunyuan3d2025hunyuan3domni,zhao2025deepmesh,bpt} require explicit point clouds as input and reconstruct the corresponding complete mesh. We therefore use the point clouds predicted by CA-VGGT as input.
The aforementioned methods fail to achieve satisfactory alignment with the input observations due to their stochastic nature, which demonstrates that our deterministic noise initialization can provide stronger control given a geometric prior.

We also compare our method with recent approaches that combine reconstruction and generation models~\cite{chang2025reconviagen, lin2026mix3rmixingfeedforwardreconstruction, li2026mvsam3d}. ReconViaGen~\cite{chang2025reconviagen} and Mix3R~\cite{lin2026mix3rmixingfeedforwardreconstruction} combine reconstruction prior by conditioning the generation process on geometry regression features~\cite{wang2025vggt, wang2026pi}, While MV-SAM3D~\cite{li2026mvsam3d} fuse conditions of each view by adaptive weighting strategies.
Our method outperforms this feature-conditioning approach by deterministically injecting reconstruction prior into the visible regions while retaining the flexibility to complete and refine the remaining part.

To further explore the full potential of our approach,
we additionally evaluate our method using a ground-truth visible point cloud as the reconstruction prior. The results in this setting provide an empirical upper bound and demonstrate that improvements in reconstruction quality directly translate into better generation performance.
Overall, our method consistently outperforms all competing approaches by a clear margin, and the advantages become more pronounced as the number of views increases. 
\paragraph{Different Number of Views.}
To evaluate robustness with respect to the number of input views, we compare our method with baseline methods that are acceptable with different numbers of views under varying view counts. We exclude LGM~\cite{tang2024lgm} and Hunyuan2.0-MV because they are designed for exactly four input views captured from predefined poses. As demonstrated in Tab.~\ref{tab:other_views}, in general, our method consistently achieves strong performance across all view counts, demonstrating its robustness to different observation settings.
In particular, compared to recent methods that combine reconstruction and generation, our advantage becomes more pronounced as the number of input views increases. We attribute this trend to the fact that cross-attention-based approaches must aggregate a growing number of conditioning features, whereas our method converts the increasingly accurate multi-view reconstruction into a fixed-size set of initial-noise tokens, allowing it to benefit more effectively from additional observations.

\begin{table}[t]
\centering
\caption{Quantitative comparison on Dorabench with baseline methods with different numbers of views. We report Chamfer Distance (CD, $\downarrow$) and F-Score ($\uparrow$, threshold=0.1). Best results are shown in \textbf{bold}. In the Type column, R, G, and C denote Reconstruction Methods, Multi-view Conditioned 3D Generative Models, and Combined Methods, respectively. $\checkmark$ indicates methods that require ground-truth camera poses.}

\resizebox{\columnwidth}{!}{%
\begin{tabular}{c@{\hspace{3pt}}l@{\hspace{3pt}}|cc|cc|cc|cc} 
\toprule
\multirow{2}{*}{Type} & \multicolumn{1}{c|}{\multirow{2}{*}{Method}} & \multicolumn{2}{c|}{2 views} & \multicolumn{2}{c|}{8 views} & \multicolumn{2}{c|}{16 views} & \multicolumn{2}{c}{24 views} \\\cline{3-10}

 & & CD $\downarrow$ & F-Score $\uparrow$ & CD $\downarrow$ & F-Score $\uparrow$ & CD $\downarrow$ & F-Score $\uparrow$ & CD $\downarrow$ & F-Score $\uparrow$\\
\midrule
R & InstantMesh~\cite{xu2024instantmesh}~$\checkmark$ & 0.083 & 0.908 & 0.070 & 0.932 & 0.072 & 0.928 & 0.154 & 0.758
\\
R & MAtCha GS~\cite{guedon2025matcha}~$\checkmark$ & 0.169 & 0.723 & 0.059 & 0.942 & 0.040 & 0.963 & 0.040 & 0.960 \\
R & DP-GS~\cite{DP-GS}~$\checkmark$ & 0.251 & 0.593 & 0.061 & 0.925 & 0.042 & 0.967 & 0.042 & 0.964 \\
\midrule
G & Trellis-MV~\cite{trellis} & 0.066 & 0.931 & 0.057 & 0.952 & 0.063 & 0.940 & 0.084 & 0.901\\
\midrule
C & ReconViaGen~\cite{chang2025reconviagen} & 0.062 & 0.952 & 0.047 & 0.971 & 0.044 & 0.977 & 0.046 & 0.974 \\
C & Mix3R~\cite{lin2026mix3rmixingfeedforwardreconstruction} & 0.068 & 0.929 & 0.054 & 0.955 & 0.054 & 0.955 & 0.054 & 0.957 \\
C & MV-SAM3D~\cite{li2026mvsam3d} & 0.070 & 0.927 & 0.055 & 0.955 & 0.056 & 0.956 & 0.056 & 0.953 \\
C & Ours & \textbf{0.056} & \textbf{0.946} & \textbf{0.031} & \textbf{0.978} & \textbf{0.027} & \textbf{0.982} & \textbf{0.029} & \textbf{0.980} \\
\bottomrule
\end{tabular}
}

\label{tab:other_views}
\end{table}

\subsection{Visualization Results}

\begin{figure*}
    \centering
    \includegraphics[width=0.75\linewidth]{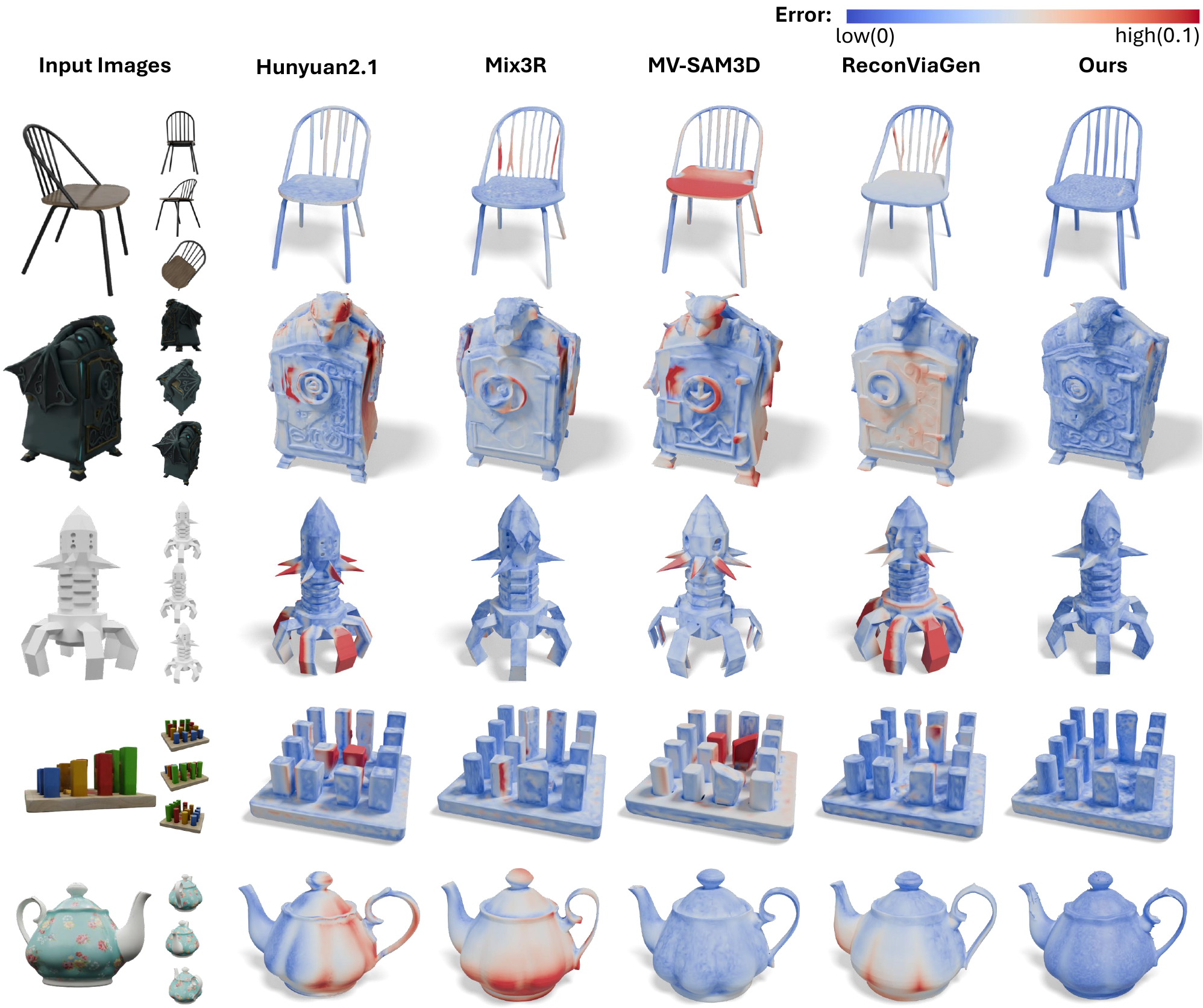}
    \caption{Qualitative comparison with baseline approaches. Per-point distance errors are color-coded to visualize geometric discrepancies. Our method achieves more faithful alignment with the input observations while better preserving fine-grained details and repeated structures. Refer to supplementary for more comparison results.}
    \label{fig:comparison}
\end{figure*}

We provide qualitative comparisons to further demonstrate the effectiveness of our approach. To visualize geometric consistency, we compute the per-point distance error and color-code the errors on the generated surfaces, clipping values to $[0,0.1]$. 
As shown in Fig.~\ref{fig:comparison}, geometry-aware initialization effectively complements the generative prior, producing shapes that align more faithfully with the input observations. The improvement is particularly evident in fine structural details and repeated patterns, such as the number of rocket support legs and the pillars on the back of the chair.

To evaluate robustness in real-world scenarios, we further test our model on in-the-wild images, as shown in Fig.~\ref{fig:teaser}. We extract the foreground objects using an existing segmentation method~\cite{liu2021paddleseg} before applying our reconstruction pipeline. The results demonstrate that our method generalizes effectively to natural out-of-domain inputs.

\begin{table}[t]
\caption{
Ablation study of the individual components. We conduct experiments using both a single-view diffusion model~\cite{hunyuan3dv21} and our multi-view diffusion model.
Noise modulation is decomposed into noise padding and noise interpolation.
In VGGT+ICP mode, we generate an anchor mesh first and align the point cloud predicted by origion VGGT~\cite{wang2025vggt} to the anchor using scaled ICP.
}
\centering
\resizebox{\columnwidth}{!}{
\footnotesize
\setlength{\tabcolsep}{12pt}

\begin{tabular}{lcc|cc} 
\toprule
\multirow{3}{*}{Method} & \multicolumn{4}{c}{Diffusion Model} \\
\cline{2-5}
 & \multicolumn{2}{c|}{Single-view Diffusion} & \multicolumn{2}{c}{Multi-view Diffusion} \\
\cline{2-5}
 & CD$\downarrow$ & F-score$\uparrow$ & CD$\downarrow$ & F-score$\uparrow$ \\
\midrule
VGGT~\cite{wang2025vggt}+ICP & 0.107 & 0.848 & 0.110 & 0.846 \\
\midrule
Random Noise & 0.084 & 0.893 & 0.060 & 0.944 \\
Random Noise(GT Cam Pose) & $-$ & $-$ & 0.059 & 0.944 \\
\textbf{w} Noise Inv & 0.070 & 0.916 & 0.056 & 0.952 \\
\textbf{w} Noise Padding & 0.049 & 0.956 & 0.048 & 0.964 \\
\textbf{w} Noise Interp & \textbf{0.045} & \textbf{0.965} & \textbf{0.042} & \textbf{0.966} \\
\bottomrule
\end{tabular}
\vspace{-10px}
}
\label{tab:ablation}
\end{table}

\subsection{Ablation Studies}
To isolate the contribution of each component, we conduct ablation studies on the same 300 Dorabench objects under the 4-view setting and report the results in Tab.~\ref{tab:ablation}. We evaluate our design with both the off-the-shelf single-view diffusion model~\cite{hunyuan3dv21} and our trained multi-view diffusion model. Experiments on single-view diffusion demonstrate that our noise-space injection mechanism can serve as a plug-and-play approach for incorporating multi-view geometric information into existing diffusion models without additional training. Our trained multi-view diffusion model provides further gains through explicit multi-view conditioning.

In both settings, directly using the unmodulated inversion noise overconstrains the diffusion process to the incomplete and imperfect CA-VGGT reconstruction. Noise padding preserves the capacity for completing unobserved regions, while confidence-aware noise interpolation reduces the influence of unreliable geometric predictions. 
Notably, noise interpolation provides a larger improvement with the multi-view diffusion model, which we attribute to the stronger guidance supplied by multi-view conditioning.

CA-VGGT resolves the coordinate mismatch between reconstruction and generation models by directly predicting point clouds in canonical space. To evaluate this design, we compare it with a post-hoc alignment baseline. Specifically, we first generate an anchor shape in canonical space from random noise and then align the point cloud reconstructed by the original VGGT to this anchor using scaled ICP. The inferior performance of this baseline highlights the importance of predicting geometry directly in canonical-space.
We further evaluate the multi-view diffusion model using both ground-truth camera poses and poses predicted by CA-VGGT to assess its robustness to small pose errors.

\section{Conclusion}

In this paper, we introduce ReconPlusGen, a novel unified framework that bridges 3D reconstruction and generation by initializing the diffusion process with partial reconstructions. Unlike existing methods, we generate deterministic initial noise from partial reconstructions and assign different range of flexibility through noise modulation, effectively shielding observed regions from the stochasticity of diffusion models while leveraging generative imagination to complete unobserved parts. 
We validated our approach through extensive experiments. Both quantitative and qualitative results demonstrate that ReconPlusGen achieves state-of-the-art performance, significantly outperforming existing baselines.

{
    \small
    \bibliographystyle{ieeenat_fullname}
    \bibliography{main}
}

\clearpage
\hypersetup{pageanchor=false}
\setcounter{page}{1}
\maketitlesupplementary

In this supplement, we first provide more experimental results in Sec.~\ref{sec:experimental_results}, including analyzes of the mapping function used in noise inversion and modulation and additional comparisons with point-cloud-conditioned methods. We then provide further details on the implementation in Sect.~\ref{sec:implementation_details}, including the CA-VGGT orientation design and a more detailed mathematical explanation of noise modulation. Finally, we discuss the strengths, limitations, and potential extensions of our method in Sec.~\ref{sec:discuss}.

\section{More Experimental Results}
\label{sec:experimental_results}

\subsection{More visualization Results}

\begin{figure*}[!t] %
    \centering
    \includegraphics[width=0.85\textwidth]{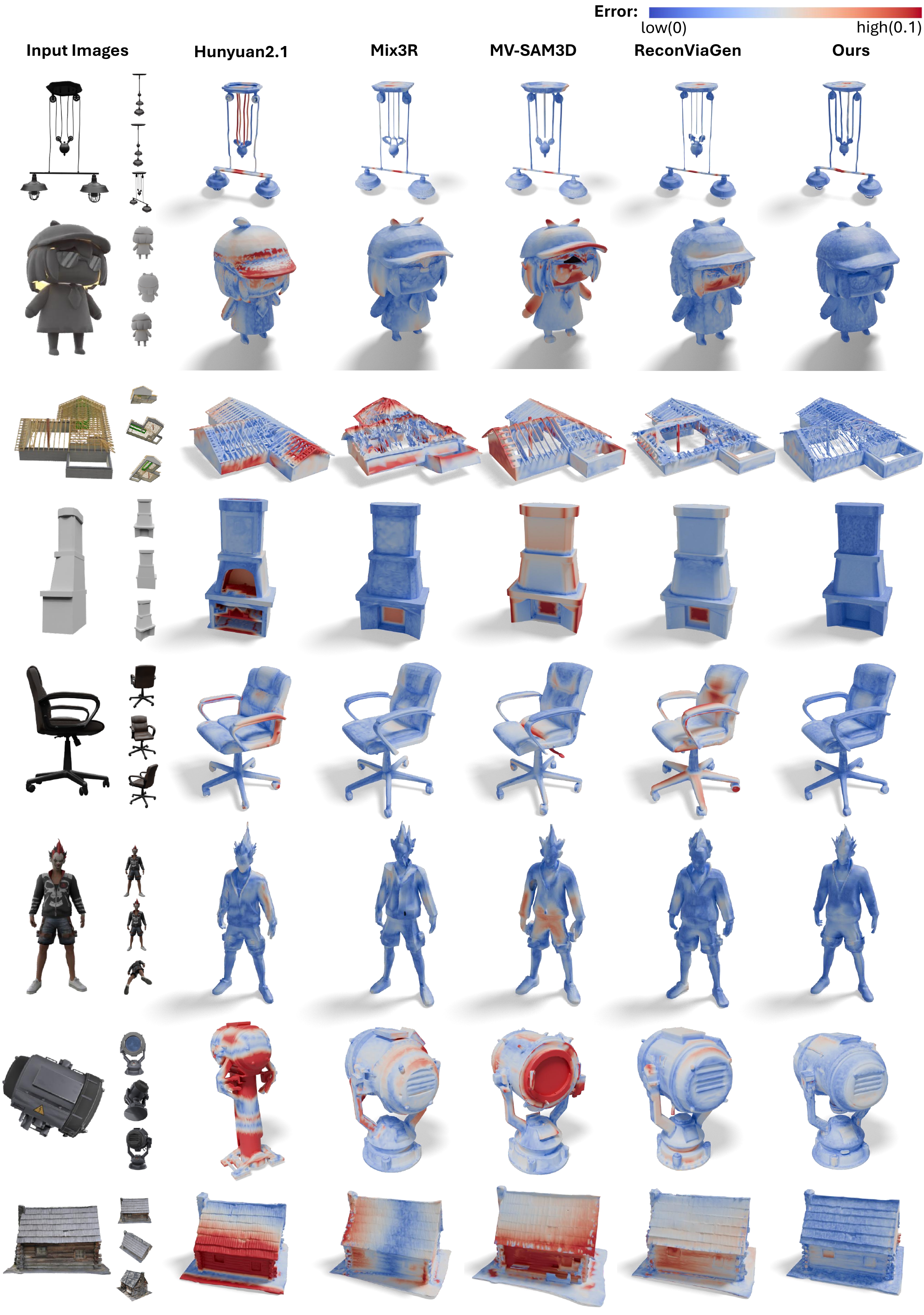} %
    \caption{More comparison results of our method and other baselines. We draw distance error with color map for visualization. } %
    \label{fig:comparisions_supp}
\end{figure*}
We report more visualization results in Fig.~\ref{fig:noise_inversion_more}. Our method consistently outperforms the baseline methods, particularly in preserving complex structures and maintaining accurate proportions among different object parts.

\subsection{Expansion Order in Noise Inversion}
\begin{figure*}[!t] %
    \centering
    \includegraphics[width=0.85\linewidth]{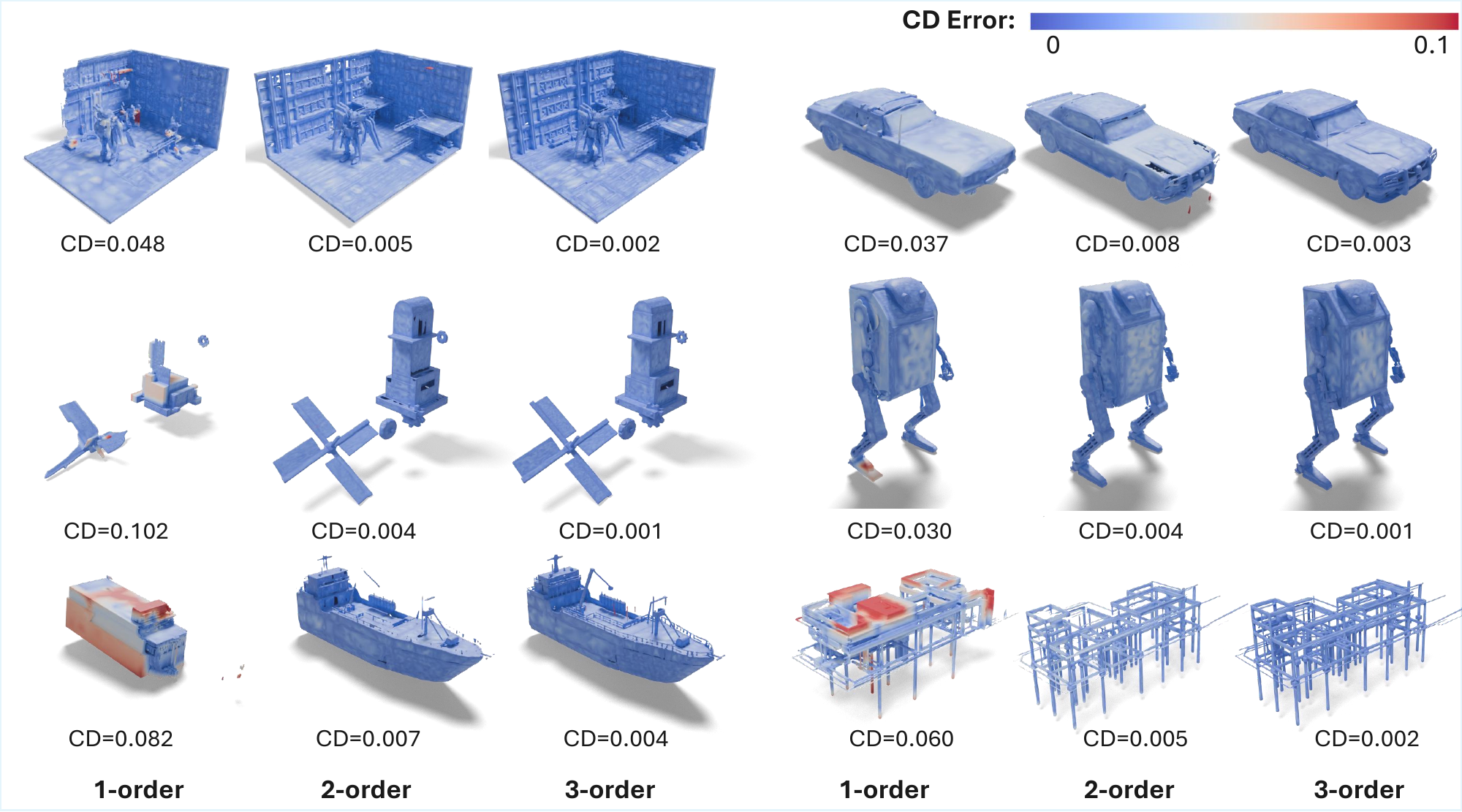} %
    \caption{Vision and numerical results of noise inversion. In most cases, noise inversion with expansion order as 2 is sufficient for geometry info injection.} %
    \label{fig:noise_inversion_more}
\end{figure*}

In Sec.~3.1 of the main paper, we initialize the diffusion process with noise obtained through inversion, where the inversion velocity is approximated using a Taylor expansion.
To visualize the effects of different expansion orders, we select several ground-truth meshes, perform noise inversion followed by denoising and decoding, and measure the errors introduced at each order. We color-code these distance errors as described in Sec.~\ref{subsec:evaluation_and_visualization} and present the results in Fig.~\ref{fig:noise_inversion_more}. 
In most cases, the error introduced by the second-order Taylor expansion is sufficiently small for our task. Note that the inversion and generation processes use the same image conditioning.

\subsection{Ablation of Noise Mapping Function in Noise Modulation}
\label{subsec:noise_modulation}

As discussed in Sec.3.4 in main context, we employ a noise modulation scheme that allows the diffusion process to rectify prediction errors from CA-VGGT. We introduce a mapping function $f(\cdot)$ that converts confidence scores into interpolation weights. Given a set of query points $q$, point cloud $S$ and confidence maps $C$, we search 15 nearest neightbors in the predicted point clouds for each query point, retrieve their corresponding confidence value, and map the NN-averaged VGGT confidence $c$ to a interpolation weight via a mapping function. We conduction ablations for several mapping functions, which is detailed in below.

\paragraph{Linear mapping.}
Each sample is first min--max normalized, then affinely rescaled to \([c_{\min},c_{\max}]\):
\begin{equation}
  \tilde{c}
  =\frac{c-c_{\mathrm{data,min}}}{c_{\mathrm{data,max}}-c_{\mathrm{data,min}}},
\end{equation}
\begin{equation}
  \alpha
  =\tilde{c}\,(c_{\max}-c_{\min})+c_{\min}.
\end{equation}
Since we fully trust the reconstructed geometry in high-confidence regions, we always set $c_{\max}=1$.
Thus, $\alpha$ spans $[c_{\min},c_{\max}]$ for each sample, independent of the absolute scale of $c$.

\paragraph{Zero-anchored sigmoid.}
In this setting, we map the raw confidence value to $[0,1]$:
\begin{equation}
\begin{aligned}
  \alpha
  &=\bigl(2\,\sigma(s\cdot c)-1\bigr)_{[0,1]}, \\
  &=\tanh\bigl(sc/2\bigr)_{c\ge 0}.
\end{aligned}
\end{equation}
where \(\sigma(z)=(1+e^{-z})^{-1}\) and \(s>0\) is a scale.
This function is smooth, monotonic, and bounded in $[0,1]$, while $s$ controls how sharply the interpolation weight responds to changes in confidence.

\paragraph{Exponential.}
\begin{equation}
  \alpha
  =1-\exp\bigl(-s\cdot c \bigr).
\end{equation}
This function increases rapidly at low confidence values and gradually saturates toward one, with $s$ controlling the rate of saturation.

We evaluate the different mapping functions under the same 4-view DoraBench dataset and report the results in Tab.~\ref{tab:mapping_func}.
At every tested scale, replacing the original linear mapping with either the zero-anchored sigmoid or exponential mapping increases the Chamfer distance and decreases the F-score.
The performance gap is smallest at \(s=0.1\), which produces the most gradual nonlinear mappings, and largest at \(s=10\).
We therefore retain the original linear mapping with \(c_{\min}=0.5\).
\begin{table}[t]
\centering
\caption{Ablation study of different mapping functions on DoraBench. We report Chamfer Distance (CD, $\downarrow$) and F-Score ($\uparrow$, threshold=0.1). Best results are shown in \textbf{bold}. }
\label{tab:mapping_func}
\resizebox{0.8\linewidth}{!}{
\begin{tabular}{llcc}
\toprule
Mapping & Scale \(s\) & CD \(\downarrow\) & F-score \(\uparrow\) \\
\midrule
Linear mapping & --- & \textbf{0.0424} & \textbf{0.9659} \\
\midrule
Sigmoid & \(0.1\) & 0.0490 & 0.9533 \\
Sigmoid & \(1\)   & 0.0607 & 0.9359 \\
Sigmoid & \(10\)  & 0.0622 & 0.9324 \\
\midrule
Exponential & \(0.1\) & 0.0512 & 0.9489 \\
Exponential & \(1\)   & 0.0616 & 0.9334 \\
Exponential & \(10\)  & 0.0623 & 0.9334 \\
\bottomrule
\end{tabular}
}
\end{table}

We further analyze the effect of $c_{\min}$ on the mapping function and report the results in Tab.~\ref{tab:noise_mapping}. As shown, $c_{\min}=0.5$ achieves the best performance. We therefore use $c_{\min}=0.5$ for all other experiments reported in the main paper.

\begin{table}[t]
\centering
\caption{Quantitative comparison on different number of $c_{min}$. We evaluate on Dorabench with 4 rendering images for each shape, and report Chamfer Distance (CD, $\downarrow$) and F-Score ($\uparrow$, threshold=0.01). We set $\delta=0.5$ by default.}
\resizebox{\linewidth}{!}{
\begin{tabular}{cc|cc|cc|cc|cc|cc} 
\toprule
\multicolumn{2}{c|}{} & \multicolumn{2}{c|}{$0.2^2$} & \multicolumn{2}{c|}{$0.4^2$} & \multicolumn{2}{c|} {$\mathbf{0.5^2}$} & \multicolumn{2}{c|}{$0.6^2$} & \multicolumn{2}{c}{$0.8^2$}\\
\cline{0-11}
 CD $\downarrow$ & F-Score $\uparrow$ & CD $\downarrow$ & F-Score $\uparrow$ & CD $\downarrow$ & F-Score $\uparrow$ & CD $\downarrow$ & F-Score $\uparrow$ & CD $\downarrow$ & F-Score $\uparrow$ & CD $\downarrow$ & F-Score $\uparrow$ \\
\midrule
{0.056} & {0.940}  & {0.052} & {0.947} & {0.048} & {0.957} & \textbf{0.042} & \textbf{0.966} & {0.046} & {0.963} & {0.047} & {0.960}\\
\bottomrule
\end{tabular}
}
\label{tab:noise_mapping}
\end{table}

\section{More Implementation Details}
\label{sec:implementation_details}
\subsection{Data Preparation}
While rendering images for training, we uniformly sampled cameras on a sphere of radius 7.5 centered at the origin and independently applied positional jitter along each axis within $[-0.2,0.2]$. For each RGB image, we also rendered the corresponding depth map. To simulate images captured by different real-world devices, we randomly varied the field of view (FOV) between $38^\circ$ and $63^\circ$. We sampled 81,920 surface points from each shape for shape encoding.

\subsection{Necessity and Design Details of CA-VGGT}

\begin{figure}[!t]
    \centering
    \includegraphics[width=1.\linewidth]{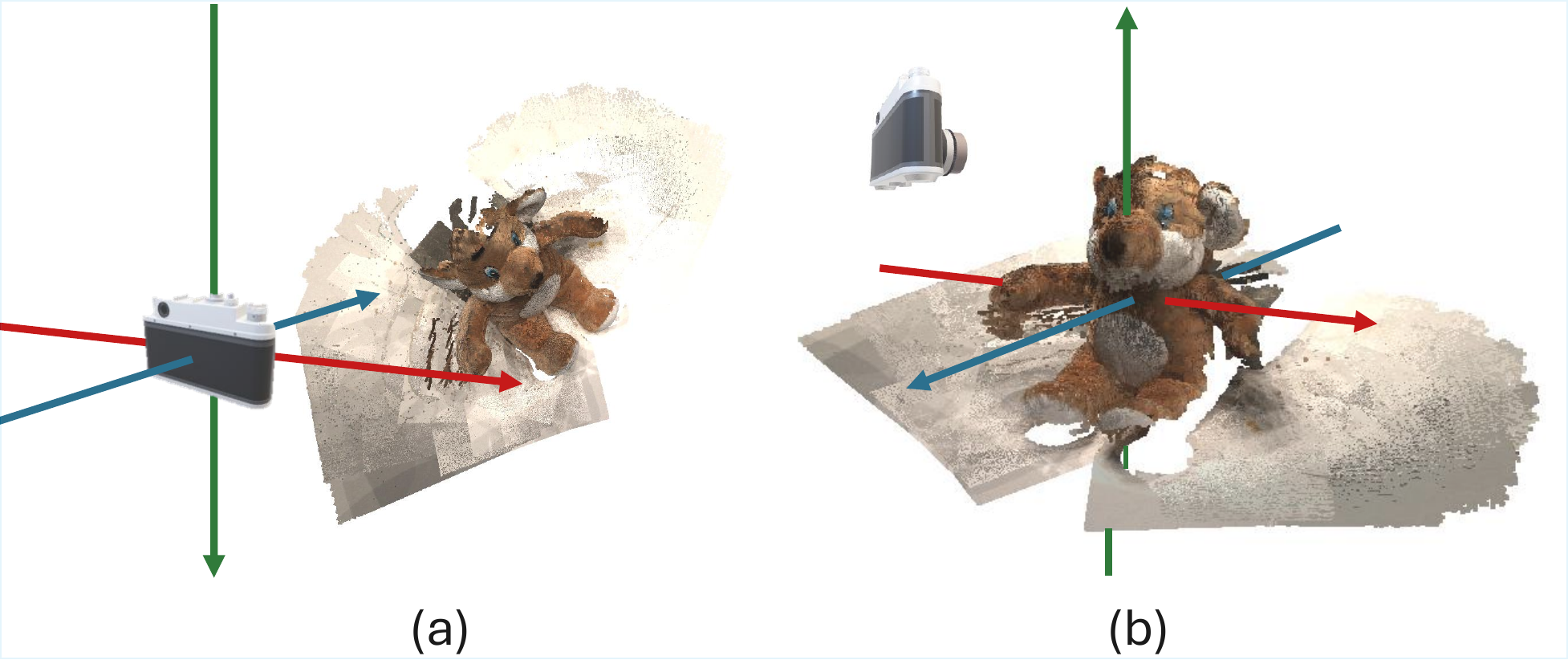}
    \caption{\textbf{Camera Coordinate.} (a) The original VGGT~\cite{wang2025vggt} represents cameras in a \emph{relative} coordinate system defined with respect to a reference view. 
(b) In contrast, our method expresses camera poses directly in a shared, object-centric \emph{canonical} coordinate system.}
\vspace{-0.5cm}
    \label{fig:coordinate}
\end{figure}
A fundamental obstacle in integrating large feed-forward 3D foundation models into 3D object generation is the \emph{coordinate system mismatch} between the two model families, as shown in Fig.~\ref{fig:coordinate}. Feed-forward 3D foundation models, such as VGGT~\cite{wang2025vggt}, typically designate the first view as the reference frame, while 3D object generation models are typically trained within a unified, object-canonical coordinate space.
This mismatch introduces systematic transformation ambiguities (rotation, scale, and axis orientation).
Existing methods either overlook this issue or rely on heuristic post-hoc alignment~\cite{chang2025reconviagen}, which cannot provide a principled and stable solution.

We present the details of our canonical space definition. Due to the ambiguity of the “front” direction for many objects and the lack of orientation annotated 3D datasets, we do not attempt to learn a model that predicts geometry in an unified object canonical space (where the object's front faces the $-Z$ axis, assuming objects are Y-up). Instead, we relax the definition of canonical space by allowing the front of the object to align with any of four possible axes: $X$, $-X$, $Z$, or $-Z$. We refer to this formulation as relative canonical space.

Specifically, during data preparation, we assume the object is placed in one of these four orientations. We then sample camera poses and render corresponding RGB and depth images accordingly. Note that the camera pose is defined with respect to a global world coordinate system, which can be aligned with the relative canonical space via a global rotation.
During training, we randomly sample $n$ images and determine the relative canonical space by collapsing the azimuth angle of the camera pose of the first image lie within $[-45^\circ, 45^\circ]$.

\subsection{Evaluation and Visualization}
\label{subsec:evaluation_and_visualization}
For the 4-view DoraBench benchmark, we use views with indices $[9,18,19,20]$. To evaluate performance under varying numbers of input views, we begin with the 2-view configuration $[9,18]$ and progressively add views to increase object coverage and reduce unobserved regions. The full 24-view configuration is $[9,\allowbreak 18,\allowbreak 19,\allowbreak 20,\allowbreak 22,\allowbreak 23,\allowbreak 24,\allowbreak 25,\allowbreak 30,\allowbreak 31,\allowbreak 34,\allowbreak 35,\allowbreak 36,\allowbreak 37,\allowbreak 38,\allowbreak 39,\allowbreak 0,\allowbreak 1,\allowbreak 2,\allowbreak 3,\allowbreak 4,\allowbreak 5,\allowbreak 6,\allowbreak 7]$.

When comparing our method with MAtCha Gaussian, we introduce a mask loss during the Gaussian splatting optimization stage with weight $0.1$ to suppress background floaters, which would otherwise produce spurious triangles in the final mesh.
For Mix3R, we evaluate the V2 implementation released on GitHub, as it supports more flexible camera configurations.
To visualize the Chamfer distance, we color each vertex of the predicted shape according to its error. For each predicted vertex, the error is defined as the maximum of two values: (1) its minimum distance to the ground-truth shape and (2) the distance from a ground-truth vertex for which it is the nearest vertex on the predicted shape.

\section{Discussion}
\label{sec:discuss}
In this paper, we propose a novel method that directly predicts shapes in canonical space and injects the predicted geometry through noise inversion and modulation. Our approach excels at preserving reconstruction fidelity by eliminating the stochasticity of diffusion models, while retaining the ability to complete unobserved regions. Despite these promising results, we identified several limitations. Specifically, we encode the predicted point clouds using an off-the-shelf Shape VAE, which is originally trained on clean, complete shapes. When the input prediction is sparse and noisy, the VAE tends to preserve these errors, leading to imperfect surfaces in the final output. Fine-tuning the Shape VAE on our VGGT predictions seems to be a clear direction for future improvement.

\end{document}